# Deep Learning-Based Classification Of the Defective Pistachios Via Deep Autoencoder Neural Networks

Mehdi Abbaszadeh
Dept. of Electrical Engineering
Vali-e-Asr University of
Rafsanjan, Iran
Email:
mahdi.abb98@yahoo.com

Aliakbar Rahimifard
Dept. of Electrical Engineering
Vali-e-Asr University of
Rafsanjan, Iran
Email:
aliakbarrahimifard@gmail.com

Mohammadali Eftekhari
Dept. of Electrical
Engineering
Vali-e-Asr University of
Rafsanjan, Iran
Email:
ma.efte.1996@gmail.com

Hossein Ghayoumi Zadeh
Dept. of Electrical
Engineering
Vali-e-Asr University of
Rafsanjan, Iran
Email:
h.ghayoumizadeh@vru.ac.ir

Ali Fayazi
Dept. of Electrical Engineering
Vali-e-Asr University of Rafsanjan, Iran
Email:
a.fayazi@vru.ac.ir

Ali Dini
Pistachio Safety Research Center
Rafsanjan University of Medical Sciences,
Rafsanjan, Iran
Email:
ali.dini2008@gmail.com

Mostafa Danaeian
Dept. of Electrical Engineering
Vali-e-Asr University of Rafsanjan, Iran
Email:
danaeian@vru.ac.ir

*Abstract*—**Pistachio nut is mainly consumed as raw, salted or roasted because of its high nutritional properties and favorable taste. Pistachio nuts with shell and kernel defects, besides not being acceptable for a consumer, are also prone to insects damage, mold decay, and aflatoxin contamination. In this research, a deep learning-based imaging algorithm was developed to improve the sorting of nuts with shell and kernel defects that indicate the risk of aflatoxin contamination, such as dark stains, oily stains, adhering hull, fungal decay and Aspergillus molds. This paper presents an unsupervised learning method to classify defective and unpleasant pistachios based on deep Auto-encoder neural networks. The testing of the designed neural network on a validation dataset showed that nuts having dark stain, oily stain or adhering hull with an accuracy of 80.3% can be distinguished from normal nuts. Due to the limited memory available in the HPC of university, the results are reasonable and justifiable.**

## I. INTRODUCTION

Dried fruits contain rich nutrients and therefore they are highly vulnerable to contamination with toxic fungi and aflatoxins due to poor weather, processing and storage conditions. Pistachio is one of the most important nuts in the world. Pistachio nuts are consumed mainly as raw, salted, and shelled because of their high nutritional properties and favorable taste. Besides, pistachio nuts are widely used in the food industry such as snack, ice cream and sweets (1). According to the Food and Agriculture Organization (2012), Iran is the world's largest producer of pistachios by producing 472097 tonnes (2). Aflatoxin, as a member of the mycotoxin family, is a secondary metabolite of molds, such as Aspergillus flavus and A.parasithicus. The fungi both on the tree and during the processing and storage stages can contaminate the product, and if the product is contaminated with their spores and provide the desired temperature and humidity conditions, the hogs grow and produce the fungus and fungi also produce poison. Although a small number of pistachios are contaminated with aflatoxin but the rate of contamination is high and a large pistachio mass will be contaminated (3).

Due to the proximity of the dimensions and characteristics of pistachio nuts, an effective separation system could not be implemented by the operator or using mechanical or electro-optical devices. Besides, the use of manpower, will reduce the accuracy and efficiency in the task of separation and increase the cost of using the vision machine in time. Therefore, the task of classification should be effectively implemented by a machine vision system with multiple processing capabilities of pistachio features (4). Machine vision techniques are increasingly used in a variety of application fields such as automatic inspection, process control, and robot guidance, usually in industry.

In recent years, Machine vision which is based on image processing is one of the modern alternative techniques that has attracted a lot of attention in recent years. Machine vision systems are proven as the most powerful method for assessing agricultural products. The artificial computerized categorization system that has been considered to simulate human decision-making for product quality has recently been heavily studied (5).

In (6), an intelligent sorting system is designed for smelly and unpolluted pistachios. The proposed system consisted of a feeder, a phonetic section, an electronic control unit, an airborne mechanism for rejecting closed-shell pistachio nuts and artificial neural networks separator. Identification is based on the PCA combination of pistachio collisions and artificial neural networks. To obtain useful specifications both time and frequency-domain were achieved. In another recent study, the machine vision system was used for egg volume estimation. The machine vision system and various methods of artificial categorizing are implemented for the division of raisins into four classes. To create a uniform illumination and eliminate ambient noise, they used a fluorescent dye at the top of the samples and selected the background for the photo portion as black (7).

Model identification has become a very important issue because of the urgent need for machine learning and artificial intelligence in practical issues. Deep learning is a hierarchical structure network that simulates the structure of the human brain to extract the features of internal and external input data. Deep learning is based on algorithms using multilayer neural network such as deep neural networks, deep belief networks, convolutional neural networks, recurrent neural networks, and Stacked Auto-encoders. These algorithms allow computers and machines to model our world well to show intelligence. In the Auto-encoders, the hidden layer provides a better display than the original raw input, and the hidden layer is always the input data density, which are important input features. Therefore, the proposal of this article is to use a deep Auto-encoder network to create a deep learning recognition system for clustering defective pistachios from healthy ones.

## II. MATERIAL AND METHODS

In this section, we introduce the SAE design for pistachio classification. Auto-encoder is an artificial neural network used to learn efficient coding. An SAE is a neural network composed of several Auto-encoder sparse layers in which the outputs of each layer are connected to the inputs of the sequential layer. The goal of an auto-encoder is to learn the compressed replay for a dataset. Auto-encoders are composed of three layers or more consisting of an input layer and a number of even smaller hidden layers that will form encryption, and ultimately an output layer, in which each neuron has the same meaning as the input layer. The hidden layer attempts to represent the input layer. In other words, auto-encoder is a function consisting mainly of two parts, an encoder section, which is a feature extraction function and calculates the feature vector from the inputs. The other part is a decoder which is actually defined probabilistic models by a specific probability function and is trained to maximize the similarity of data in which the output units are directly connected to the input units, as shown in Figure 1 (8).

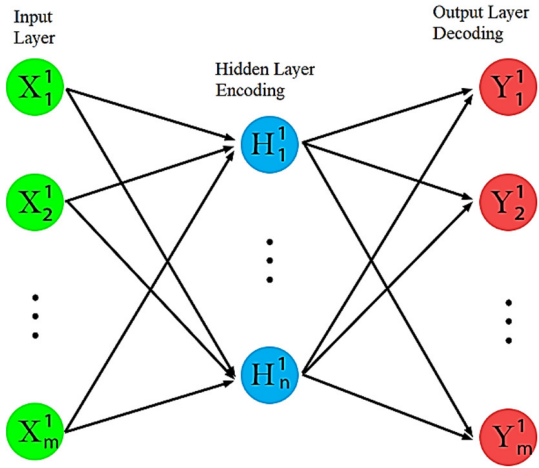

Figure 1: Structure of the sparse auto-encoder

The proposed sparse auto-encoder algorithms were trained on the input of $X_n^l$, the hidden layer of $H_m^l$, and the output layer of $Y_n^l$, where n is the number of input or output neurons, m is the number of hidden neurons and l is the number of sparse auto-encoder. The output layer maps the input vector $I_n^l$ to the hidden layer $H_m^l$ with a non-linear function s.

$$H_m^l = s(\sum_{i=0}^{n}(w_i * X_i^l) + b_m) \tag{1}$$

Where $w_i$ is the parameters (or weights) associated with the connection between the input unit and the hidden unit. $b_m$ are Biases in the hidden layer. s(v) is the sigmoid function. The sigmoid function is defined as below:

$$s(v) = \frac{1}{1+e^{-v}} \tag{2}$$

The output layer $Y_n^l$ has the same number of units with the input layer and is defined as:

$$y_n^l = s(\sum_{j=0}^{m}(\widehat{w}_j * H_j^l) + b_n) \tag{3}$$

Where $\widehat{w}_j$ are the parameters (or weights) associated with the connection between the hidden unit and the output unit. $b_m$ are Biases in the hidden layer. s is the sigmoid function in equation. We introduce Stacked Auto-encoders design for pistachio classification. The first sparse Auto-encoder contains input layer to learn primary features $H_m^1$ raw input which is shown in Figure 2 (8).

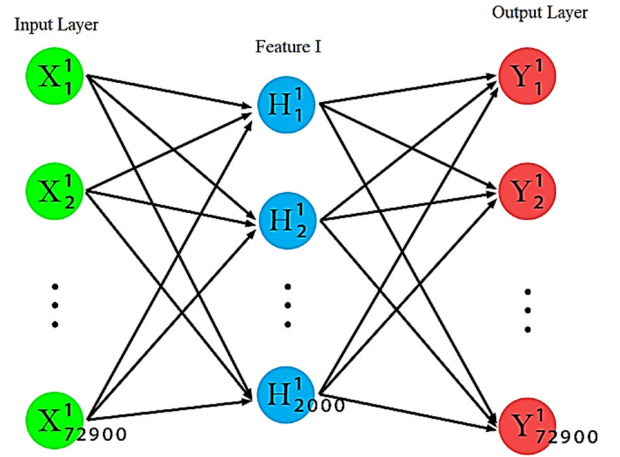

Figure 2: The first sparse Auto-encoder

The first sparse auto-encoder generates the main feature (feature I). The main feature presents $H_m^1$ input layer to the second sparse auto-encoder that generates secondary features (II). In Figure 3, the primary features that are used as raw inputs to the next generation of Auto-encoder for using secondary features are shown.

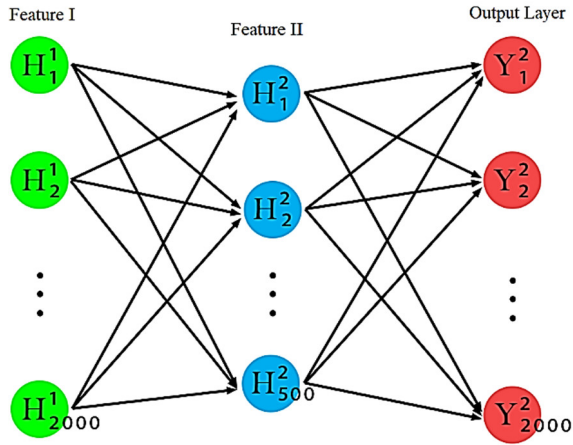

Figure 3: The second sparse-encoder

Then, the secondary feature behaves as the input layer for a softmax classifier for mapping the secondary features to the digit labels shown in Figure 4.

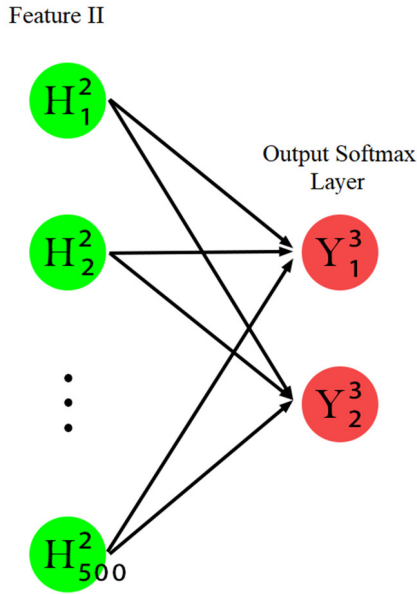

Figure4: Softmax classifier

Finally, the first and second sparse auto-encoder was combined with the softmax classifier to produce three layers of autoencoder, as shown in Figure 5.

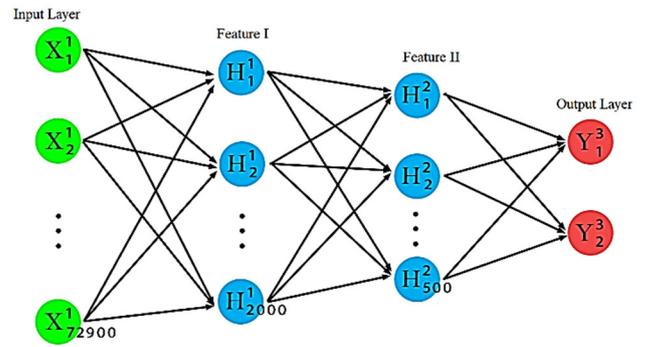

Figure 5: The proposed design of Auto-encoder architecture

Stacked Auto-encoders have two hidden layers (primary and secondary features) and an output layer (softmax classifier) that are able to classify pistachios.

### III. RESULTS

The total number of collected pistachio images was 305, of which 214 images are related to defective pistachios and 91 images are related to healthy pistachios. The images were first converted to the gray-level format and then enhanced before being inputted to the neural network. An example of the neural network input images is shown in Figure 6.

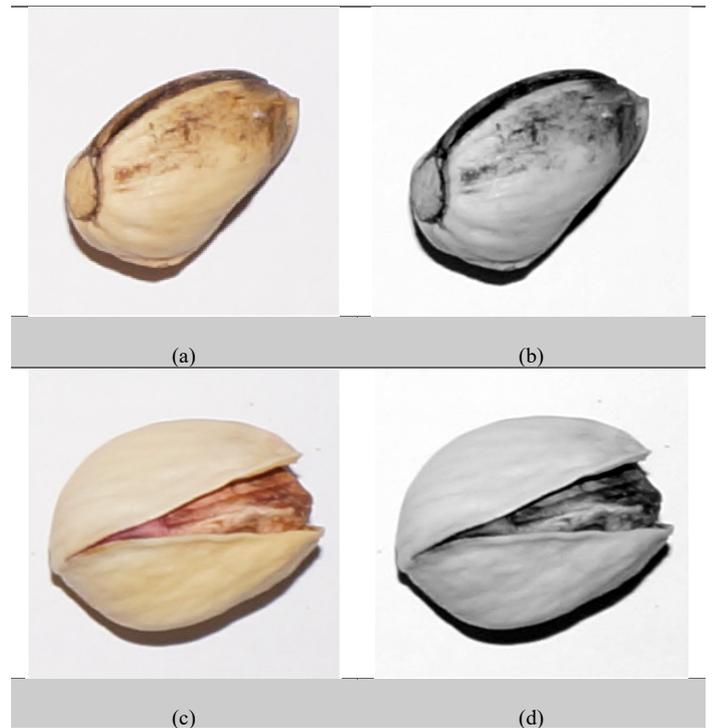

Figure 6. (a)Pistachio Nut Aflatoxin in the format of RGB (b) Pistachio Nut Aflatoxin in the format of Graylevel(c) Healthy pistachios in the format of RGB (d) Healthy pistachios in the format of Graylevel

In this section, we evaluated the performance of the stacked auto-encoders. The stack autoencoder included an encoder with input layers of 72,900. The auto-encoder discovered how to learn the features of every 270*270 pixels of the image. Experiments are implemented in MATLAB 2018 programming environment. The auto-encoder is performed using the MATLAB deep learning toolbox. The first sparse autoencoder with an input layer of 72900, a hidden layer of 2000 and an output layer of 72900 is shown in Figure 7. The first sparse autoencoder is trained to produce 2000 primary features. The second layer of sparse autoencoder was designed with an input layer of 2000, a hidden layer with a size of 196 (half the input) and an output layer of 2000, as shown in Figure 8. The second sparse autoencoder is trained to produce 500 secondary features.

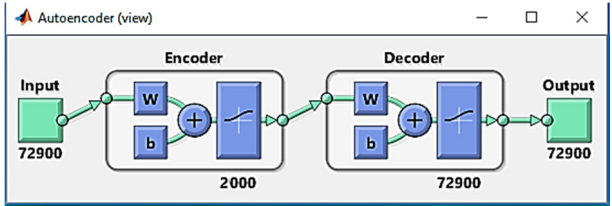
Figure 7. the proposed first sparse autoencoder

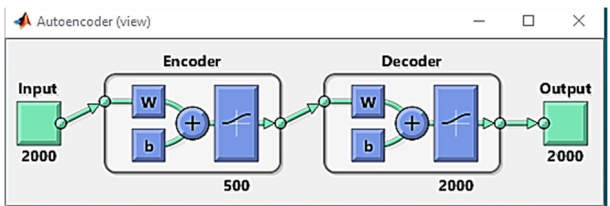
Figure 8. The proposed second sparse autoencoder

The first and second sparse autoencoder uses regularization L2 to learn sparse display. Regularization controls the effect of a regularizer L2 for network weight (and not bias). Finally, these 500 features feed the softmax layer (figure 4). The softmax layer is trained to produce two output classes. In Figure 9, the proposed design of stack autoencoder is represented by an input layer of 72900, hidden layers with 2000 primary features, 500 secondary features, and an output layer with 2 labels. The proposed stack autoencoder shown in Figure 9 teaches 305 test data from the image.

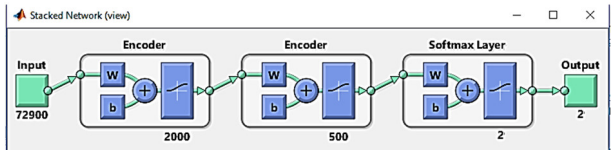
Figure 9. The proposed stacked autoencoder

To create better results, after finishing the training phase, fine-tuning using back-propagation can be done to improve the results by setting the parameters of all layers at a time. We feed our autoencoder with 305 tutorials images. The mapping learned by the encoder of autoencoder can be useful for extracting data properties. Each neuron in the encoder has a weighted vector shown in Figure 10.

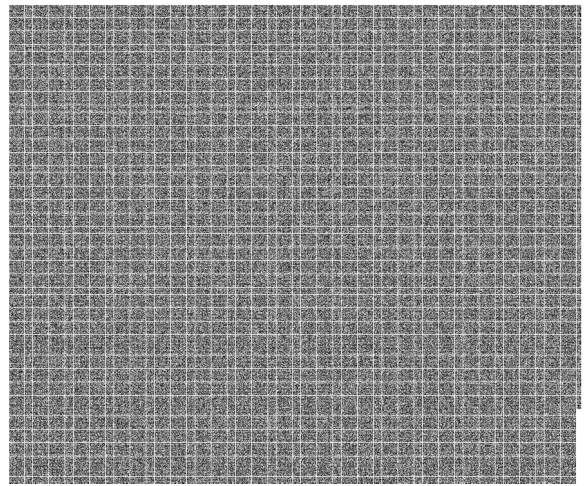
Figure 10. Visualizing the weights of the first autoencoder

IV. DISCUSSION

Confusion matrix of data images is shown in Figures 11. The first two diagonal cells in the confusion matrix represent the number and percentage of correct classifications by the trained network.

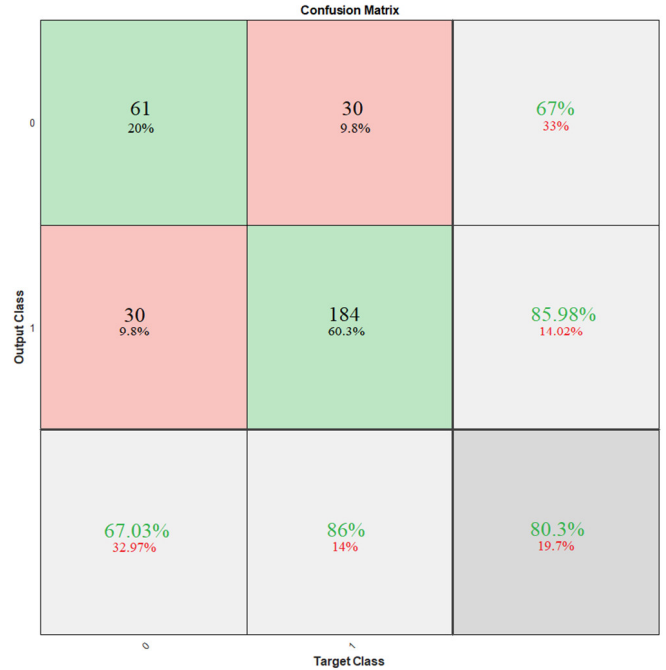
Figure 11. The confusion matrix of proposed stacked autoencoder

The plot of training progress is shown in figure 12.

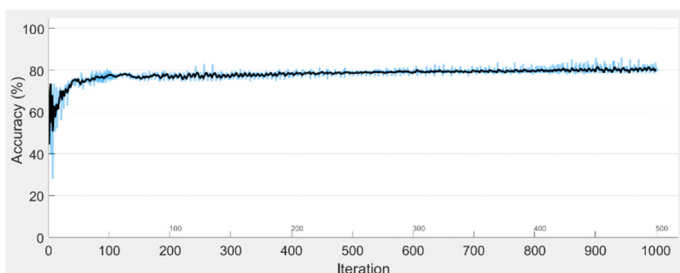

Figure 12. Training progress according to the accuracy and itration

Due to the limited memory available in HPC of university, the results are reasonable and justifiable. By increasing the number of images, the accuracy of the method will be much higher.

## V. Conclusion

Pistachios, either hand-picked or picked up mechanically, should be dried and classified after peeling. The primary separation of low-quality pistachios is usually by floating them in water and using mechanical separators. Imaging methods have been used as a potential tool for quick and non-invasive evaluation and quantification of toxic fungi and aflatoxins in fruit and dried fruits. The visual machine system, using neural network algorithms in image processing, can perform product classification operations by analyzing images immediately. The most developed methods for isolating and quickly obtaining information on infected fruits are the methods used to investigate aflatoxin.

Artificial neural network is a suitable model for biological processing that describes the application of complex mathematical approaches. In this paper, we showed the effectiveness of deep stacked learning to classify healthy and defective pistachios. This study is described the possibility of using image classification until pistachio with oily stains, dark stains, adhering hull (Likely poor core quality) is separated from normal nuts with a credit of false positive rate of 9.8% and a false negative rate of 9.8%.

In the real-time application, as the final purpose for the auto-detection of pollution and the high efficiency of fruit classification, is still in the early stages of research. Therefore, more study is needed to improve the model's strength, detection limit, and sorting efficiency for fast detection. To further improve the accuracy, reliability, and speed of the recognition of these approaches, especially for online evaluation, the development of new and effective chemometrics algorithms is essential. Also, combining other techniques can reduce the drawbacks of a single technique, which can be another way. In particular, due to the diversity of sample volumes and shapes, it is very difficult to simultaneously collect contamination information on the entire core surface to determine the speed. Changing the components of other factors (For example, temperature and humidity) can interfere with the accuracy of the diagnosis. Finally, it's very difficult to exclude contaminated specimens accurately using existing devices (For example air nozzles, mechanical arm) due to the limited precision of hardware integration.

## Acknowledgment

We are really thankful for the contribution of the Food and Drug department of Rafsanjan University of Medical Sciences that have accompanied us to collect the database.

## References

[1] Kola O, Hayoğlu İ, Türkoğlu H, Parıldı E, Ak BE, Akkaya MR. Physical and chemical properties of some pistachio varieties (Pistacia vera L.) and oils grown under irrigated and non-irrigated conditions in Turkey. Quality Assurance and Safety of Crops & Foods. 2018:1-6.

[2] Gupta V, Khare K, Singh R, editors. Fpga design and implementation issues of artificial neural network based pid controllers. Advances in Recent Technologies in Communication and Computing, 2009 ARTCom'09 International Conference on; 2009: IEEE.

[3] Bond TC, Chang A, Zhou J. Real-time, in-situ detection of volatile profiles for the prevention of aflatoxin fungal contamination in pistachios. Lawrence Livermore National Lab.(LLNL), Livermore, CA (United States); 2017.

[4] Omid M, Firouz MS, Nouri-Ahmadabadi H, Mohtasebi SS. Classification of peeled pistachio kernels using computer vision and color features. Engineering in Agriculture, Environment and Food. 2017;10(4):259-65.

[5] Shahabi M, Rafiee S, Mohtasebi SS, Hosseinpour S. Image analysis and green tea color change kinetics during thin-layer drying. Food Science and Technology International. 2014;20(6):465-76.

[6] Kavdır I, Guyer D. Evaluation of different pattern recognition techniques for apple sorting. Biosystems Engineering. 2008;99(2):211-9.

[7] Mollazade K, Omid M, Arefi A. Comparing data mining classifiers for grading raisins based on visual features. Computers and electronics in agriculture. 2012;84:124-31.

[8] Loey M, El-Sawy A, EL-Bakry H. Deep Learning Autoencoder Approach for Handwritten Arabic Digits Recognition. arXiv preprint arXiv:170606720. 2017.